%% file: 0main.tex
\documentclass{article}

\usepackage{microtype}
\usepackage{graphicx}
\usepackage{subcaption}
\usepackage{booktabs} 

\usepackage{hyperref}

\usepackage[accepted]{icml2026}

\usepackage{amsmath}
\usepackage{amssymb}
\usepackage{mathtools}
\usepackage{amsthm}
\usepackage{xspace}
\usepackage{pifont}
\usepackage{adjustbox}
\usepackage{multirow}
\usepackage{bigstrut}

\usepackage[capitalize,noabbrev]{cleveref}

\theoremstyle{plain}

\theoremstyle{definition}

\theoremstyle{remark}

\usepackage[textsize=tiny]{todonotes}

\icmltitlerunning{\ours: Mitigating Hallucinations via Recurrent Cross-Layer Reasoning in Large Vision-Language Models}
\newcommand{\ours}{\texttt{HalluRNN}\xspace}

\makeatletter
\renewcommand{\ICML@appearing}{}
\makeatother
\begin{document}

\twocolumn[
  \icmltitle{\ours: Mitigating Hallucinations via Recurrent Cross-Layer Reasoning in Large Vision-Language Models}

  \icmlsetsymbol{equal}{*}

  \begin{icmlauthorlist}
    \icmlauthor{Le Yu}{equal}
    \icmlauthor{Kaishen Wang}{equal}
    \icmlauthor{Jianlong Xiong}{}
    \icmlauthor{Yue Cao}{}
    \icmlauthor{Lei Zhang}{}
    \icmlauthor{Zhang Yi}{}
    \icmlauthor{Tao He}{}
  \end{icmlauthorlist}

  \icmlcorrespondingauthor{Tao He}{tao\_he@scu.edu.cn}

  \icmlkeywords{Machine Learning, Vision-Language Models, Hallucination}

  \vskip 0.3in
]

\printAffiliationsAndNotice{\icmlEqualContribution}
\begin{abstract}
Though Large Vision-Language Models (LVLMs) have achieved remarkable performance across various tasks, they are still prone to hallucinations—generating outputs that are textually plausible but visually ungrounded. While prior approaches generally address this issue through data-centric fine-tuning or innovative decoding strategies, these methods often require substantial resources or task-specific configurations. In this work, we introduce an architecture-level solution, \ours, which enhances model stability through recurrent cross-layer reasoning. Specifically, we propose a novel Dual-Gated Depth Propagation Unit (DG-DPU) module, which is shared across layers and recurrently refines hidden states. This allows for the adaptive propagation of information throughout the model, enforces consistency across layers, and mitigates hallucinations caused by representational drift. By fine-tuning only the DG-DPU module, \ours achieves strong and robust performance across multiple benchmarks.  The code is available \href{https://anonymous.4open.science/r/HalluRNN-62EB}{here}.
\end{abstract}

\section{Introduction}


Large Vision-Language Models (LVLMs) \citep{liu2023visual,zhou2023infmllm,zhu2023minigpt} have achieved remarkable success in generating coherent text and performing multimodal tasks by integrating visual inputs with Large Language Models (LLMs) \citep{mann2020language,touvron2023llama,grattafiori2024llama}. Despite their impressive capabilities, LVLMs are still prone to hallucinations—outputs that are factually inconsistent with the given input. For example, LVLMs may describe an image containing two people as if it contains three \citep{wangdamo2025}, indicating a failure in visual grounding. Such issues can undermine user trust and pose significant challenges to real-world deployment \citep{hu2023advancing,liu2023llm,chen2024driving,huang2025survey}. Several studies have investigated the causes of hallucinations in LVLMs, attributing them to a range of factors, including the inherently generative nature of LLMs \citep{banerjee2024llms, huang2025survey}, biased or insufficient training data \citep{hu2023ciem, sahoo2024addressing}, modality misalignment \citep{li2023evaluating, wang2024mitigating}, and other related issues.

To mitigate these issues, various approaches have been proposed. Fine-tuning-based methods \citep{zhou2024aligning, yu2024rlhf} leverage high-quality data to fine-tune LVLMs, improving modality alignment and generation fidelity. Contrastive decoding (CD) strategies \citep{leng2024mitigating, zhang2024debiasing} reduce hallucinations by contrasting original logits with those obtained from intentionally distorted inputs, suppressing dominant modality-specific biases. While these end-to-end strategies have shown promise, they often overlook that hallucinations tend to emerge progressively during multi-step reasoning. Consequently, recent works have shifted toward layer-wise analysis for a deeper understanding. For example, DeCO \citep{wang2025mllm} adaptively selects relevant preceding layers and proportionally integrates their outputs into the final layer. DAMO \citep{wangdamo2025} finds that hallucinations primarily emerge in later stages due to reasoning inconsistencies or representational drift, introducing momentum-based activation accumulation to stabilize outputs during inference.




\begin{figure*}
    \centering    \includegraphics[width=0.9\textwidth]{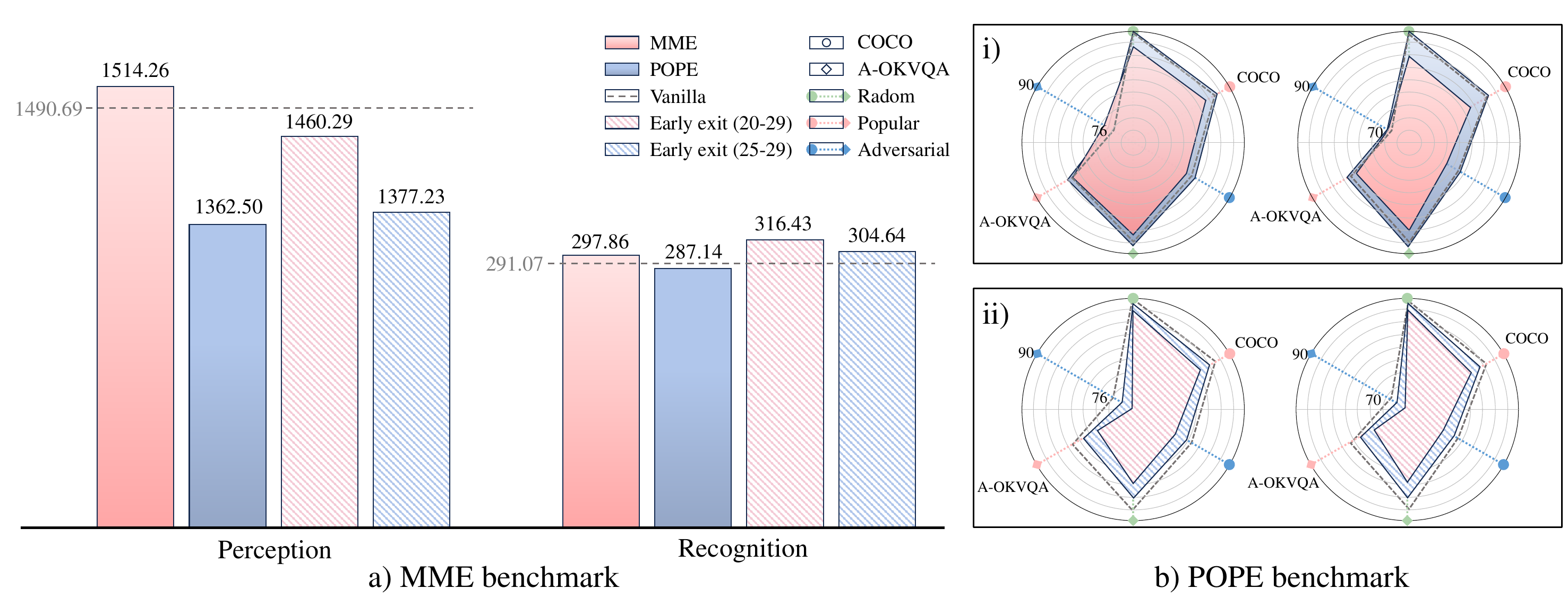}
    \caption{The performance of DAMO and DeCO under different hyperparameter settings is compared on two benchmarks using LLaVA-1.5, with the Vanilla setting results also provided for reference. (a) Comparison on MME benchmark. The red and blue bars represent DAMO with the MME setting and POPE setting, respectively, along with DeCO under different early exit configurations. (b) Comparison on POPE benchmark. Subfigure $i$ displays DAMO's performance for accuracy (left) and F1 score (right). Subfigure $ii$ illustrates DeCO's performance for accuracy (left) and F1 score (right).}
    \label{fig:layer_wise_sensitive}
\end{figure*}

However, despite their innovative analyses, existing layer-wise methods still share several limitations: \ding{182} Their inter-layer connections are relatively rudimentary, often relying on fixed-weighted accumulation or basic skip connections without any dynamic adaptation. \ding{183} These methods are highly sensitive to hyperparameter choices, frequently requiring task-specific tuning, which undermines their robustness and generalizability. Specifically, Figure \ref{fig:layer_wise_sensitive} shows the hyperparameter sensitivity of DAMO and DeCO across MME \citep{fu2023mme} and POPE \citep{li2023evaluating}  benchmarks. DAMO performs well with MME-specific hyperparameters but drops 128.19 points on MME when using POPE's settings. Similarly, adjusting DeCO's early-exit parameters significantly degrades performance on both benchmarks. While current LVLMs rarely explore sophisticated inter-layer mechanisms, the importance of cross-layer interactions has long been recognized in deep convolutional neural networks (DCNNs). From early architectures like ResNet \citep{he2016deep} to advanced strategies such as MRLA \citep{fang2023cross} and DLA \citep{wang2024strengthening}, dynamically modeling dependencies across layers has proven effective in enhancing model capacity. This observation raises a natural question: \textbf{\textit{Can hallucinations in LVLMs be mitigated through adaptive cross-layer interaction?}}

To answer this question, we propose \ours, a novel architecture enabling LVLMs to self-correct through cross-layer reasoning. The key component of \ours is the Dual-Gated Depth Propagation Unit (DG-DPU), a recurrent module shared across all Transformer layers. While originally designed to capture temporal dependencies, recurrent neural networks (RNNs) are also well-suited for modeling hierarchical relations along the depth of a network. By treating each layer as a sequential step, the DG-DPU refines and propagates hidden states throughout the model. This recurrent cross-layer reasoning mechanism enables adaptive information integration across layers, effectively mitigating hallucinations caused by reasoning inconsistencies or representational drift during inference. Notably, \ours fine-tunes only the DG-DPU module with minimal data, making it both efficient and effective. To the best of our knowledge, this is the first work to incorporate RNN-based cross-layer reasoning into LVLMs for hallucination mitigation. 

The paper’s contributions are summarized as follows:
\begin{itemize}
    \item We propose a novel framework, \ours, that integrates a new RNN module across Transformer layers in LVLMs, enabling adaptive cross-layer interaction to enhance representation consistency and effectively reduce hallucinations.
    \item We design a specialized RNN cell, the Dual-Gated Depth Propagation Unit (DG-DPU), tailored for LVLMs, which efficiently aggregates and propagates information from preceding layers, offering a lightweight and generalizable solution for stabilizing model behavior with minimal task-specific supervision.
\end{itemize}

\section{Related Work}
\paragraph{Hallucinations in LVLMs} Hallucination issue in Large Language Models (LLMs) has been extensively explored in \citet{xu2024hallucination, huang2025survey}, highlighting instances where generated responses deviate from the given input or factual knowledge. While in Large Vision-Language Models (LVLMs), the hallucination issue refers to models generating text responses that are inconsistent with the visual input \citep{leng2024mitigating,wangdamo2025}. This issue is caused by various factors, such as misalignment between textual and visual modalities \citep{wang2024mitigating}, insufficient training or fine-tuning \citep{sahoo2024addressing}, inherently generative nature of LLMs \citep{huang2025survey}, and limited attention to contextual grounding \citep{huang2024opera, zhu2024ibd}. These challenges are prevalent in various tasks such as image captioning and visual question answering (VQA).

\paragraph{Addressing Hallucinations in LVLMs}
Numerous studies have explored mitigating hallucinations in LVLMs through fine-tuning. While effective, fine-tuning is resource-intensive and depends on high-quality datasets \citep{yu2024makes} and may harm performance on other tasks \citep{barnett2024fine}. Recent works \citep{yu2024rlhf, zhou2024calibrated} have proposed aligning textual and visual modalities using RLHF, achieving significant reductions in hallucinations. With the success of DoLA \citep{chuang2023dola}, several contrastive decoding (CD) variants have emerged, such as VCD \citep{leng2024mitigating}, VDD \citep{zhang2024debiasing}, HIO \citep{chen2024alleviating}, and IBD \citep{zhu2024ibd}, which tackle hallucinations by contrasting outputs or enhancing token distinction. CODE \citep{kim2024code} uses self-generated descriptions to correct hallucinations. SID \citep{huo2025selfintrospective} adaptively amplifies and subtracts hallucinations based on visual information. OPERA \citep{huang2024opera} adds a penalty during beam search decoding. DeGF \citep{zhang2025self} verifies and corrects hallucinations using feedback from text-to-image models. Details of these methods is provided in Appendix.

Meanwhile, layer-wise correlation methods such as DAMO \citep{huang2024opera} and DeCO further address hallucinations by accumulating activations and integrating knowledge across layers. These works show that strengthening layer interactions can also help reduce hallucinations. However, despite their plug-and-play and training-free flexibility, both DAMO and DeCO rely heavily on hyperparameter selection, which limits their practical applications.

\paragraph{Layer Interaction Methods}
Layer interaction has been widely applied in Deep Convolutional Networks (DCNNs) and Transformer-based models to enhance representational ability. ResNet \citep{he2016deep} facilitates information flow through skip connections, while DenseNet \citep{huang2017densely} promotes feature reuse by concatenating outputs from all previous layers. DIANet \citep{huang2020dianet} further improves layer interaction by incorporating a parameter-shared LSTM block across layers. Additionally, various methods \citep{fang2023cross,chen2024zer0,wang2024strengthening, xiao2025muddformer,wang2025imagent} have been introduced to refine layer interaction. Notably, although DAMO \citep{wangdamo2025} was originally designed to mitigate hallucinations in LVLMs, it also functions as a layer interaction mechanism by accumulating momentum activations across all layers.

\section{Method}

\begin{figure}
    \centering
    \includegraphics[width=0.5\textwidth]{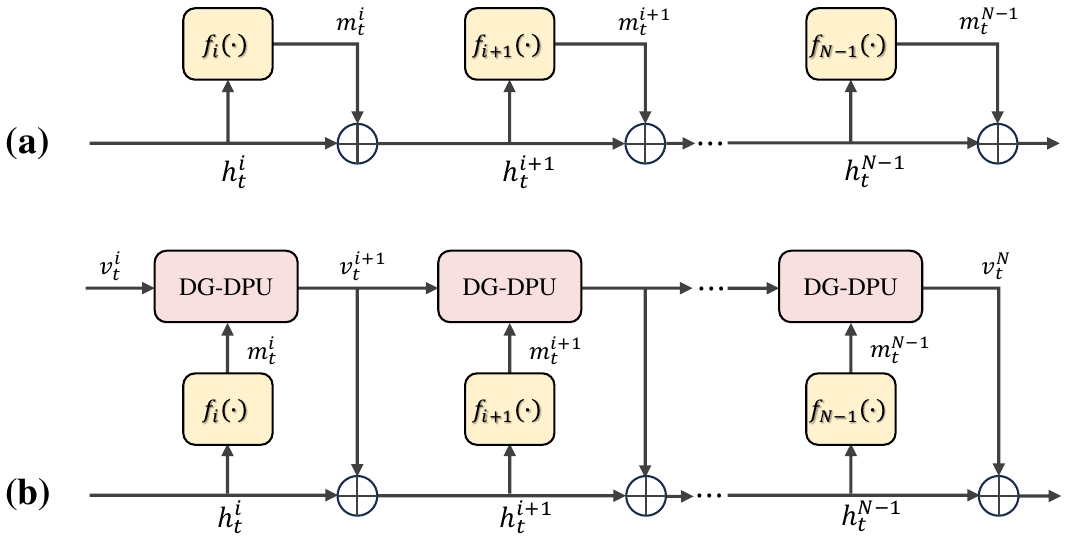}
    \caption{(a) Vanilla decoding process of an $N$-layer LVLM. (b) Proposed \ours architecture, featuring a Dual-Gate Dynamic Propagation Unit (DG-DPU) that enforces layer-wise consistency via shared gating weights.}
    \label{fig:overall_arch}
\end{figure}
\subsection{Preliminary}

\paragraph{LVLMs Architecture} 
Large Vision-Language Models (LVLMs) integrate a visual encoder $\mathcal{V}$ and a language model $\mathcal{L}$. The visual encoder extracts spatial-semantic features $\mathbf{v} = \mathcal{V}(I) \in \mathbb{R}^{d_v}$ from the input image $I$, while the language model processes a text prompt $x$ into token embeddings $\mathbf{x} = \mathcal{E}(x) \in \mathbb{R}^{d_l}$ based on a vocabulary set $\mathcal{E}$. After aligning these two modalities—vision and language—using mechanisms such as the Q-former \citep{li2023blip} or a projection layer \citep{liu2023visual}, LVLMs perform autoregressive decoding through a stack of \(N\) layers, as shown in Figure \ref{fig:overall_arch} (a). The standard decoding procedure predicts the next-token distribution based on the final-layer hidden state, formulated as follows \citep{liu2023visual,chiang2023vicuna}:
\begin{equation}
    \begin{cases} 
        m_t^{i} = f_i(h_{t}^{i}) \\
        h_{t}^{i+1} = h_{t}^{i} + m_t^{i} 
    \end{cases} 
    \quad \Rightarrow \quad 
    p(x_{t+1} | x_{:t}) = \text{softmax}(\phi(h^N_t))
\end{equation}

where $i \in \{0, \ldots, N - 1\}$ and $h_{t}^{i}$ denotes the hidden state at the $i$-th Transformer layer for the \(t\)-th token. The function $f_i(\cdot)$ refers to the $i$-th Transformer layer, and $m_t^{i}$ denotes the accumulated hidden state learned by $f_i(\cdot)$, which includes the activations generated by both the MLP and the self-attention modules. The function $\phi(\cdot)$ denotes the prediction head used to generate the next token distribution. 

\begin{figure*}
    \centering
    \includegraphics[width=0.9\textwidth]{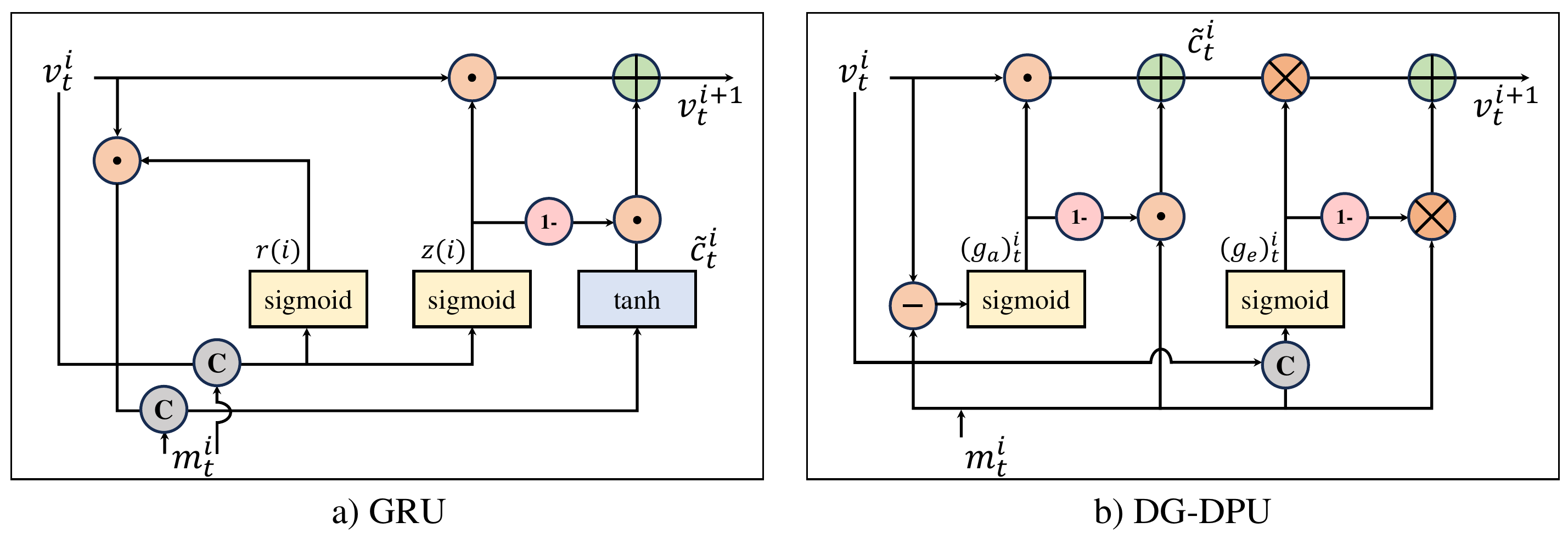}
    \caption{(a) The GRU block. (b) The proposed DG-DPU block. The symbols $\oplus$, $\circleddash$ and $\copyright$ represent matrix addition, subtraction and concatenation, respectively. Specifically, $\odot$ denotes Hadamard product, while $\otimes$ represents scalar multiplication.}
    \label{fig:gru_dg_dpu}
\end{figure*}

\paragraph{RNNs Architecture} Recurrent Neural Networks (RNNs) are designed to model sequential data by maintaining a dynamic ``memory'' of past inputs through recurrent connections. At each step, the hidden state $s^{i+1}$ depends on the current input $x^i$ and the previous hidden state $s^{i}$:
\begin{equation}\label{equ:rnn}
    s^{i+1} = \text{RNN}(x^i, s^{i}) = f(\boldsymbol{W}x^i + \boldsymbol{U}s^{i} + b)
\end{equation} 
The use of shared weight matrices $\boldsymbol{W}$ and $\boldsymbol{U}$ across time steps ensures temporal consistency and reduces the number of parameters. RNNs have demonstrated strong performance in various fields such as natural language processing (NLP) \citep{yin2017comparative, nasir2021fake}, speech recognition \citep{saon2021advancing, oruh2022long}, and time series prediction \citep{fu2022temporal, fan2023parallel}. However, traditional RNNs often struggle with long-term dependencies due to vanishing or exploding gradients. Long Short-Term Memory (LSTM) \citep{hochreiter1997long} addresses this issue by introducing gating mechanisms. Gated Recurrent Units (GRU) \citep{cho2014properties} build upon this concept while offering a simpler alternative with comparable effectiveness.

\subsection{Motivation}

Recent works such as DoLA \citep{chuang2023dola}, DeCO \citep{wang2025mllm}, and DAMO \citep{wangdamo2025} have shown that hallucination in LVLMs can be effectively mitigated through layer-wise interaction. However, these methods typically regulate inter-layer behavior through indirect mechanisms. For example, DoLA smooths predictions via contrastive decoding, while DeCO and DAMO reuse historical context or intermediate features to stabilize output representations. Although effective to some extent, these approaches lack a direct mechanism for explicitly modeling the evolution of hidden states between layers, which is central to the causes of hallucinations.

Building on the success of RNNs in capturing temporal dependencies, we propose a novel perspective: \textit{treating the sequence of hidden states across Transformer layers as a temporal process}. By introducing a lightweight RNN module across layers, we enable the model to dynamically integrate and refine information as it flows through depth. This approach explicitly models inter-layer dependencies, allowing the network to adaptively correct inconsistencies and suppress noise.

\subsection{\ours: Recurrent Cross-Layer Reasoning} 
To enhance inference stability and mitigate hallucinations in LVLMs, we propose \ours, a Recurrent Cross-Layer Reasoning architecture, which explicitly models hidden state evolution across layers via RNN-based interactions. In Figure \ref{fig:overall_arch} (b), we treat hidden state transitions along depth as a temporal process and incrementally refines the output of each Transformer layer through recurrent feedback. We formalize the inter-layer transition as:
\begin{equation}
    m_t^i = h_t^{i+1}-h_t^{i} \quad \text{for } i = 0, \ldots, N-1
\end{equation}
where $m_t^i$ captures the transition dynamics between adjacent layers, revealing how internal states evolve across depth. To regulate this evolution, we introduce a novel RNN block, Dual-Gated Depth Propagation Unit (DG-DPU) (detailed in Section \ref{sec:RNN}) that is shared across all layers, allowing the model to learn a generalized recurrence pattern over depth. 

At each layer $i$, the DG-DPU receives two inputs: the current inter-layer transition $m_t^i$ and the previous recurrent hidden state $v_t^{i}$. Following Equation \ref{equ:rnn}, it computes:
\begin{equation}
    v_t^{i+1} = \text{DG-DPU}(m_t^i, v_t^{i}) \quad
\end{equation}
This recurrent state $v_t^{i+1}$ acts as a dynamic memory that accumulates and filters the evolving information across layers. Finally, we reintegrate $v_t^{i+1}$ into the model by updating the hidden state as:  
\begin{equation}
    h_t^{i+1} = h_t^{i} + v_t^{i+1} \quad
\end{equation}
This additive update ensures that cross-layer reasoning is continuously propagated throughout the network, fostering both depth-wise consistency and stable output generation.


\subsection{Dual-Gated Depth Propagation Unit}
\label{sec:RNN}
To guide hidden state evolution across Transformer layers, we propose a Dual-Gated Depth Propagation Unit (DG-DPU) (shown in Figure \ref{fig:gru_dg_dpu} (b)). Inspired by Gated Recurrent Unit (GRU) (shown in Figure \ref{fig:gru_dg_dpu} (a)), DG-DPU incorporates specialized gating mechanisms to control information flow. Unlike standard GRU, our design introduces two custom gates—Constraint Gate and Correction Gate—tailored to the inference dynamics of LVLMs. These gates operate in a coarse-to-fine manner to mitigate hallucinations. 

{\renewcommand{\arraystretch}{0.9}
\setlength{\tabcolsep}{9pt}
\begin{table*}[h]
    \centering    
    \small
    \caption{Experimental results of various decoding strategies on the SEEM-annotated MSCOCO, A-OKVQA, and GQA datasets from POPE benchmark utilizing LLaVA-1.5 model.}
    \adjustbox{width=1\textwidth}{
        \begin{tabular}{cccccccc}
    \toprule
    \multicolumn{2}{c}{Dataset} & \multicolumn{2}{c}{MSCOCO} & \multicolumn{2}{c}{A-OKVQA} & \multicolumn{2}{c}{GQA} \\
    \midrule
    Setting  & \multicolumn{1}{c}{Decoding} & Accuracy & F1 score & Accuracy & F1 score & Accuracy & F1 score \\
    \midrule
    \multirow{9}[2]{*}{Random} & Vanilla & 89.70  & 89.79  & 87.33  & 88.52  & 86.83 & 88.11 \\
           & VCD& 87.83  & 88.13&84.80&86.37 & 84.57 & 86.33  \\
          & VDD & 89.43 & 89.41 & 87.90 &88.89 & 87.70 & 88.71 \\
          & DOLA  &89.70&89.77&87.37&88.55 &  87.07 & 88.29  \\
           & POVID  &89.50 &89.56 &87.30 &88.49 & 88.63 & 89.45\\
           & DAMO &90.03 &89.99 &87.93 &88.96 & 87.70 & 88.76 \\
           & OPERA &89.73  &89.83  & 87.20 &88.43 & 86.87 & 88.13  \\
           & DeCO & 88.87 & 89.32& 84.67 & 86.53 & 84.83 & 86.69 \\
           & \ours & \textbf{90.20} & \textbf{90.05} & \textbf{89.27} & \textbf{90.01} & \textbf{88.67} & \textbf{89.49}  \\
    \midrule
    \multirow{9}[2]{*}{Popular}  & Vanilla & 86.17  & 86.75  &80.30&83.21 & 74.60 & 79.34  \\
          &VCD&84.27 &85.10 &77.33 &80.90 & 71.60 & 77.29  \\
          & VDD &  86.43 & 86.76 & 81.67 & 84.09 & 76.03 & 80.13 \\
          & DOLA  &86.33&86.87&80.47&83.33 & 74.73  & 79.42   \\
          & POVID  &86.10 &86.63 &80.40 &83.28 & 75.93 & 80.02 \\
          & DAMO &86.70 &87.07 & 81.37&83.92 & 76.17 & 80.30 \\
          & OPERA & 86.30 &86.88  &80.07  &83.07 & 74.57 & 79.31  \\
          & DeCO & 84.73 & 85.92 & 77.47& 81.39 & 71.13 & 77.39\\
          & \ours & \textbf{87.27} & \textbf{87.45} & \textbf{81.97} & \textbf{84.29} & \textbf{76.60} & \textbf{80.49}\\
    \midrule
    \multirow{9}[2]{*}{Adversarial} & Vanilla & 79.70  & 81.70  & 69.37  & 76.12 & 68.43 & 75.55  \\
          & VCD   & 78.13  & 80.28  &67.80&74.95 & 67.70 & 74.99  \\
          & VDD &  80.76 & 82.27 & 71.07 & 77.01 & 70.33 & 76.54  \\
          & DOLA  &79.77&81.71&69.53&76.22 & 68.47 & 75.57  \\
          & POVID  &79.60 &81.53&69.83 &76.39 & 70.40 & 76.51 \\
          & DAMO & 80.43&82.08 &70.73 &76.87 & 70.03 & 76.42 \\
          & OPERA & 79.77 &81.77  & 69.07 & 75.97 & 68.40 & 75.52 \\
          & DeCO & 78.40 & 81.17 & 67.03  & 74.93  & 66.13 & 74.47 \\
           & \ours & \textbf{81.17} & \textbf{82.49} & \textbf{71.43} & \textbf{77.20} & \textbf{70.80} & \textbf{76.78} \\
    \bottomrule
    \end{tabular}%
    }\label{Resul:POPE}
\end{table*}
}
\textbf{Constraint Gate} This gate performs a coarse-grained estimation of potential hallucinations by analyzing the deviation introduced at the current layer. It identifies representational drift between adjacent layers and generates a stabilized intermediate state. The mechanism is defined as:
\begin{equation}
\left\{
\begin{aligned}
&(g_a)_t^i = \sigma(W_a \cdot (m_t^i - v_t^{i})) \\
&\tilde{c}_t^i = (g_a)_t^i \cdot v_t^{i} + (1 - (g_a)_t^i) \cdot m_t^i
\end{aligned}
\right.
\label{equa:ga}
\end{equation}
where $W_a \in \mathbb{R}^{d_l \times d_l}$ is a learnable weight matrix and $\sigma$ denotes the sigmoid activation function. By computing the element-wise difference $\Delta = m_t^i - v_t^{i}$, this gate captures abrupt deviations introduced by the current layer. These changes serve as a heuristic signal to estimate whether hallucinations might have emerged. By fusing $m_t^i$ and $v_t^{i}$, the Constraint Gate acts as a coarse filter that favors more stable representations, suppressing potential deviations. The resulting state $\tilde{c}_t^i$ serves as a preliminary estimate for further refinement.

\begin{table*}[ht]
    \centering
    \begin{minipage}[t]{0.60\textwidth}  
        \centering
        \small
        \caption{Experimental results of various decoding strategies on MM-Vet benchmark utilizing LLaVA-1.5 model.}
        \resizebox{\textwidth}{!}{ 
        \begin{tabular}{lccccccc}
            \toprule
            Decoding & Rec & Ocr & Know & Gen & Spat & Math & Total\\ \midrule
            Vanilla & 36.1 & 23.0 & 18.0 & 22.2 & 25.1 & 11.5 & 31.1  \\
            VCD & 36.1 & 22.4 & 21.0  & 23.1  & 28.4 & 3.8 & 31.1 \\
            VDD & 37.1 & 22.8 & 19.0 & 21.7 & 28.3 & 11.2 & 31.8\\
            DOLA & 36.5 & 22.3 & 18.1  & 23.0  & 25.7 & 7.7 & 30.8 \\
            POVID & 35.1  & 23.2 & 17.0  & 19.5  & 28.0 & 11.5 & 31.4 \\
            DAMO & 37.9 & 22.4 & 21.0  & 24.2  & 25.9 & 7.7 & 32.2 \\
            OPERA & 32.7 & 19.7 & 14.0 & 16.4 & 22.4 & 7.7 & 28.3 \\
            DeCO & 35.8 & 26.8 & 19.2 & 21.3 & 30.2 & 7.7 & 32.4  \\
            \ours &  36.2  & 27.1  & 18.7 & 21.7 & 31.6 &  7.7& \textbf{33.0} \\
            \bottomrule
        \end{tabular}
        }\label{Resul:MM-Vet}
    \end{minipage}
    \hspace{0.03\textwidth}  
    \begin{minipage}[t]{0.35\textwidth}  
        \centering
        \small
        \caption{Experimental results on CHAIR dataset using LLaVA-1.5.}
        \resizebox{\textwidth}{!}{ 
        \begin{tabular}{lcc}
            \toprule
             \textbf{Decoding} & CHAIR\textsubscript{\textit{s}} $\downarrow$ & CHAIR\textsubscript{\textit{i}} $\downarrow$ \\ \midrule
             Vanilla & 50.8 & 14.2  \\
             VCD & 53.4 & 15.3  \\
             VDD & 54.6 & 16.0 \\
             DOLA & 52.2 & 15.0 \\
             POVID & 50.8  & 14.7 \\
             DAMO & 53.6 & 14.9 \\
             OPERA & 52.8 & 14.6 \\
             DeCO & 42.4 & 11.6 \\
             \ours & \textbf{33.4} & \textbf{11.2} \\
            \bottomrule
        \end{tabular}
        }\label{Resul:CHAIR}
    \end{minipage}
\end{table*}
\begin{table*}[htbp]
  \centering
  \caption{Experimental results of various decoding strategies on MMMU benchmark utilizing LLaVA-1.5 model.}
  \small
    \begin{tabular}{lccccccccc}
    \toprule
    \textbf{Decoding} & \multicolumn{1}{c}{Vanilla} & \multicolumn{1}{c}{VCD} & \multicolumn{1}{c}{VDD} & \multicolumn{1}{c}{DOLA} & \multicolumn{1}{c}{POVID} & \multicolumn{1}{c}{DAMO} & \multicolumn{1}{c}{OPERA} & \multicolumn{1}{c}{DeCO} &  \multicolumn{1}{c}{\ours}  \\
    \midrule
    \textbf{Score} & 34.2  & 34.2 & 34.2  & 34.1  &    33.7   &  34.4   & 33.5 & 32.3 & \textbf{34.6} \\
    \bottomrule
    \end{tabular}
  \label{Resul:mmmu}
\end{table*}%

\textbf{Correction Gate} This gate performs fine-grained adjustments by adaptively blending the stabilized historical representation $\tilde{c}_t^i$ with the current-layer update $m_t^i$. It can be represented as follows:
\begin{equation}
\left\{
\begin{aligned}
&(g_e)_t^i = \sigma(W_{e2} \cdot \text{ReLU}(W_{e1} \cdot [m_t^i, v_t^{i}])) \\
&v_t^{i+1} = (g_e)_t^i \cdot m_t^i + (1 - (g_e)_t^i) \cdot \tilde{c}_t^i
\end{aligned}
\right.
\label{equa:ge}
\end{equation}
Here, $W_{e1} \in \mathbb{R}^{2d_l \times d_l}$ and $W_{e2} \in \mathbb{R}^{d_l \times 1}$ are learnable parameters. The concatenated input $[m_t^i, v_t^{i}]$ is passed through a two-layer projection to produce the gating scalar $(g_e)_t^i \in [0, 1]$, which adaptively modulates the integration of current and historical information.

Unlike the Constraint Gate, which provides a coarse-grained estimate of representational drift, the Correction Gate further assesses whether the current layer's update is trustworthy. For instance, when the Constraint Gate flags a high deviation, it may indicate a corrective shift rather than an error. The Correction Gate, therefore, operates at a finer granularity to decide whether to prioritize new evidence or rely on stabilized context, enabling more reliable control of cross-layer reasoning.

\section{Experiments}

\subsection{Experimental Setup}\label{sec:setup}
\paragraph{Training Details}

Following previous fine-tuning studies, we employ the LLaVA-1.5 7B architecture as our visual-language backbone. To train our proposed DG-DPU module, we leverage the specific dataset configuration from \citet{zhou2024aligning}, which comprises a subset of 17K examples randomly sampled from the LLaVA-Instruct-150 dataset \citep{liu2023visual}. During the optimization phase, we adopt a parameter-efficient strategy by exclusively fine-tuning the DG-DPU module while keeping the pre-trained backbone parameters frozen. We adhere to the training setup of \citet{zhou2024aligning} by fine-tuning for a total of 3 epochs. The training is conducted on 4 RTX 4090 GPUs, taking approximately 3 hours. More details are in Appendix. 

\begin{figure*}
	\centering
	\begin{minipage}{0.36\textwidth}
	{ 			\includegraphics[width=\textwidth]{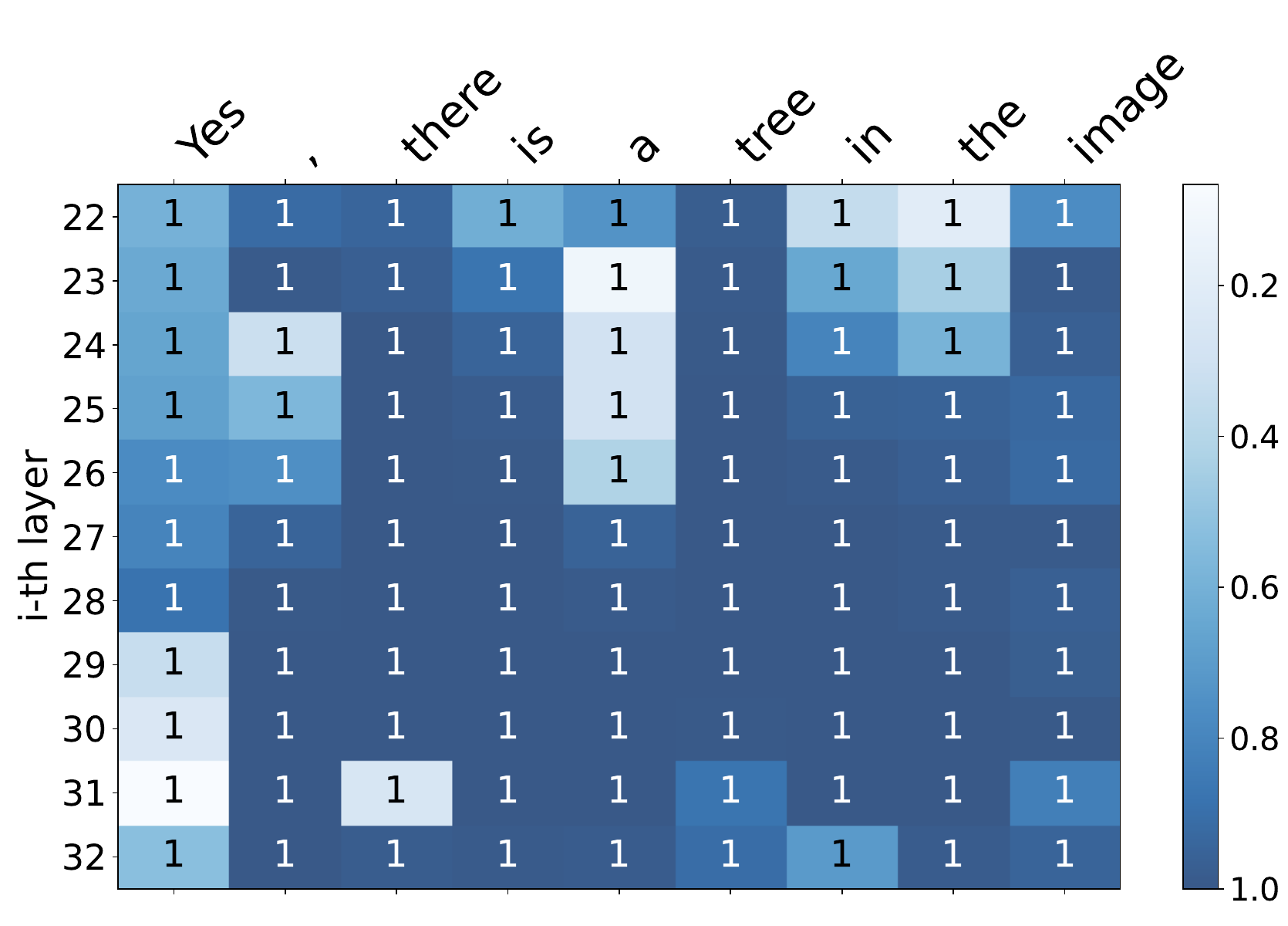}}\subcaption{Vanilla}
	\end{minipage}
	\begin{minipage}{0.36\textwidth}
	{ 			\includegraphics[width=\textwidth]{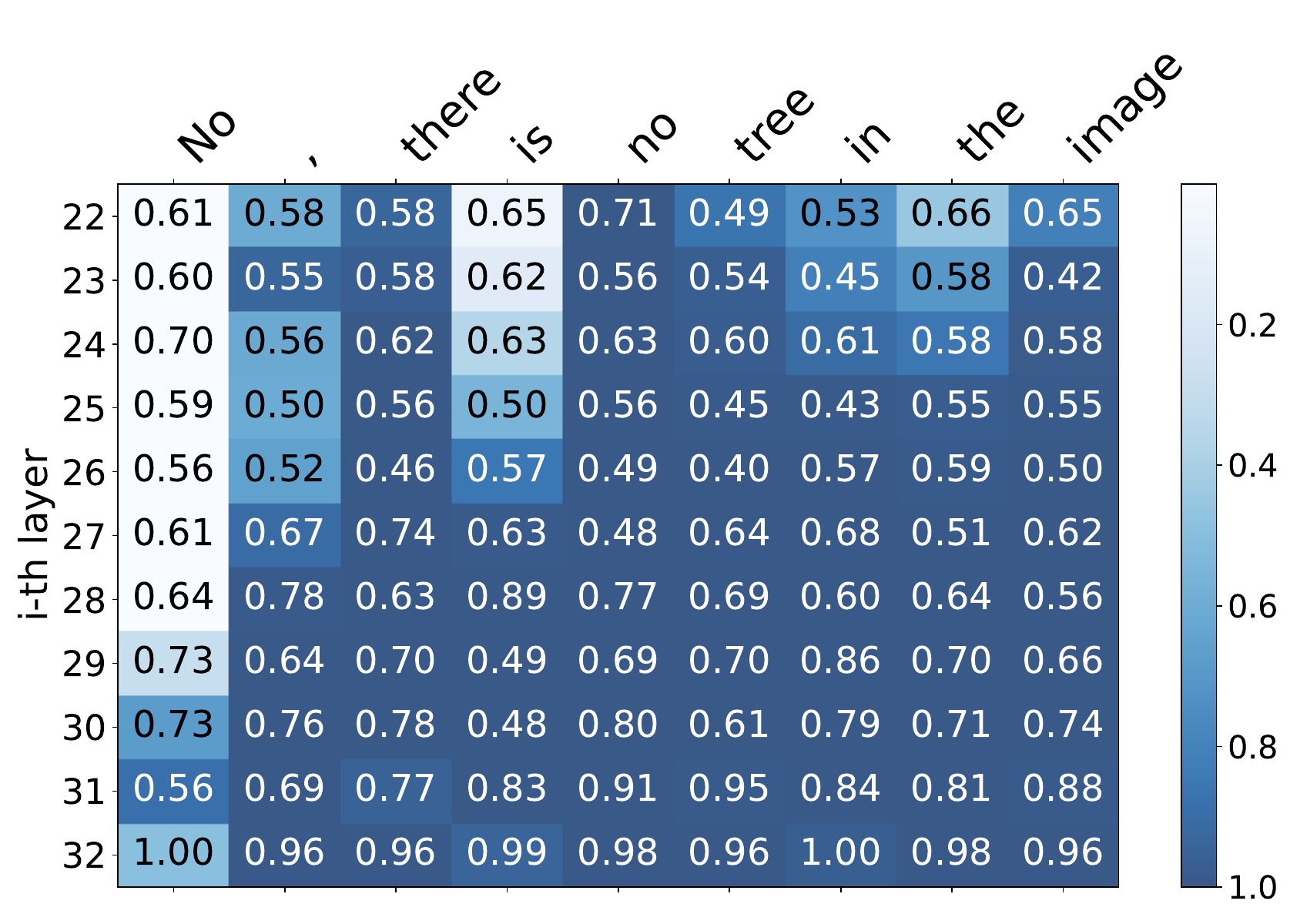}}\subcaption{\ours}
	\end{minipage}
	\begin{minipage}{0.25\textwidth}
	{ 			\includegraphics[width=\textwidth]{Figures/1/5_COCO_val2014_000000036494.jpg}}
	\end{minipage}    \caption{The visualization compares the Vanilla and \ours decoding methods in addressing the question posed by the textual input \textit{``Is there a tree in the image?''}, alongside the corresponding image from the COCO 2014 dataset. The color density of each grid indicates the probability of the output word at the $i$-th layer, while the displayed values represent the $g_e$ values, which show a clear upward trend as the network depth increases. $g_e$=1 denotes vanilla setting in Equation \ref{equa:ge}. }
\label{fig:detailed_analysis}
\end{figure*}

\begin{figure*}[h]
    \centering	
    \begin{minipage}{0.32\textwidth}
    \includegraphics[width=\textwidth]{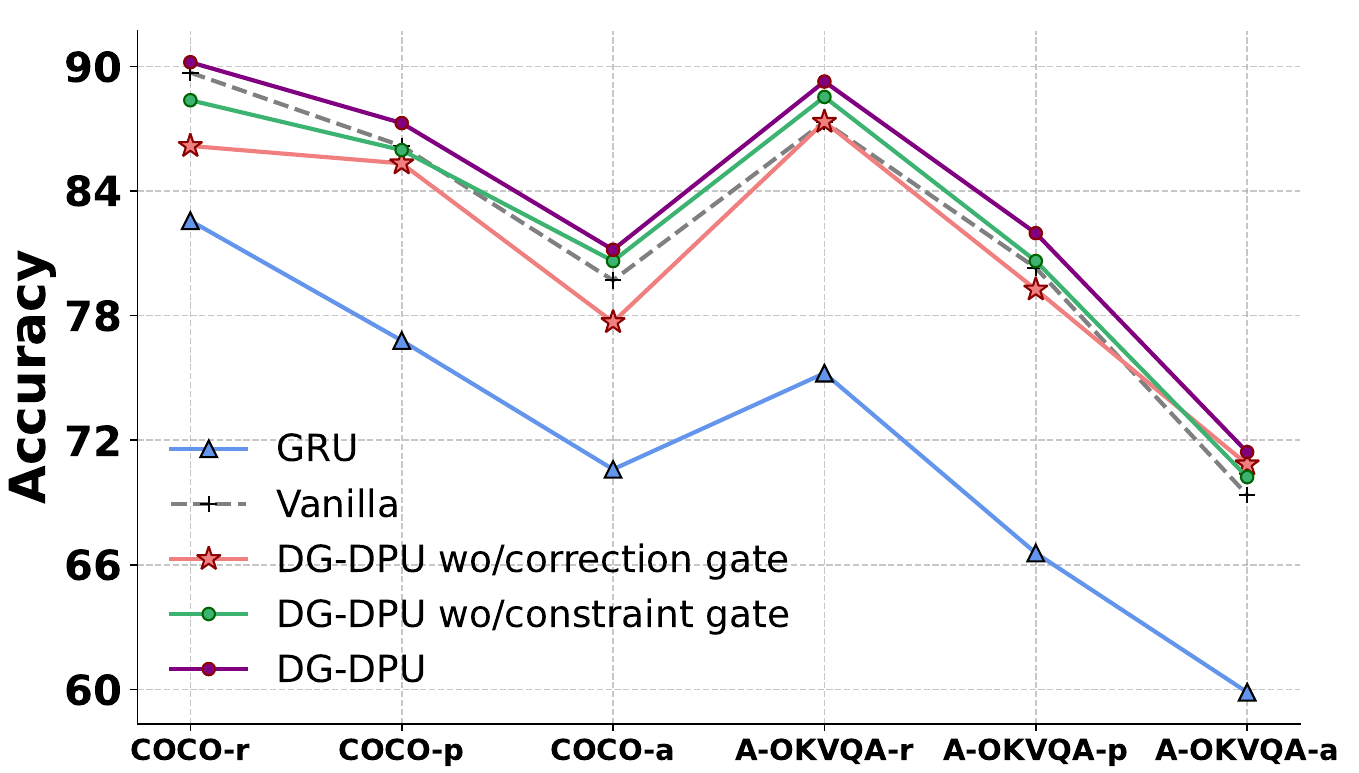}\subcaption{Acc. comparison}
    \end{minipage}
    \begin{minipage}{0.32\textwidth}
    	\includegraphics[width=\textwidth]{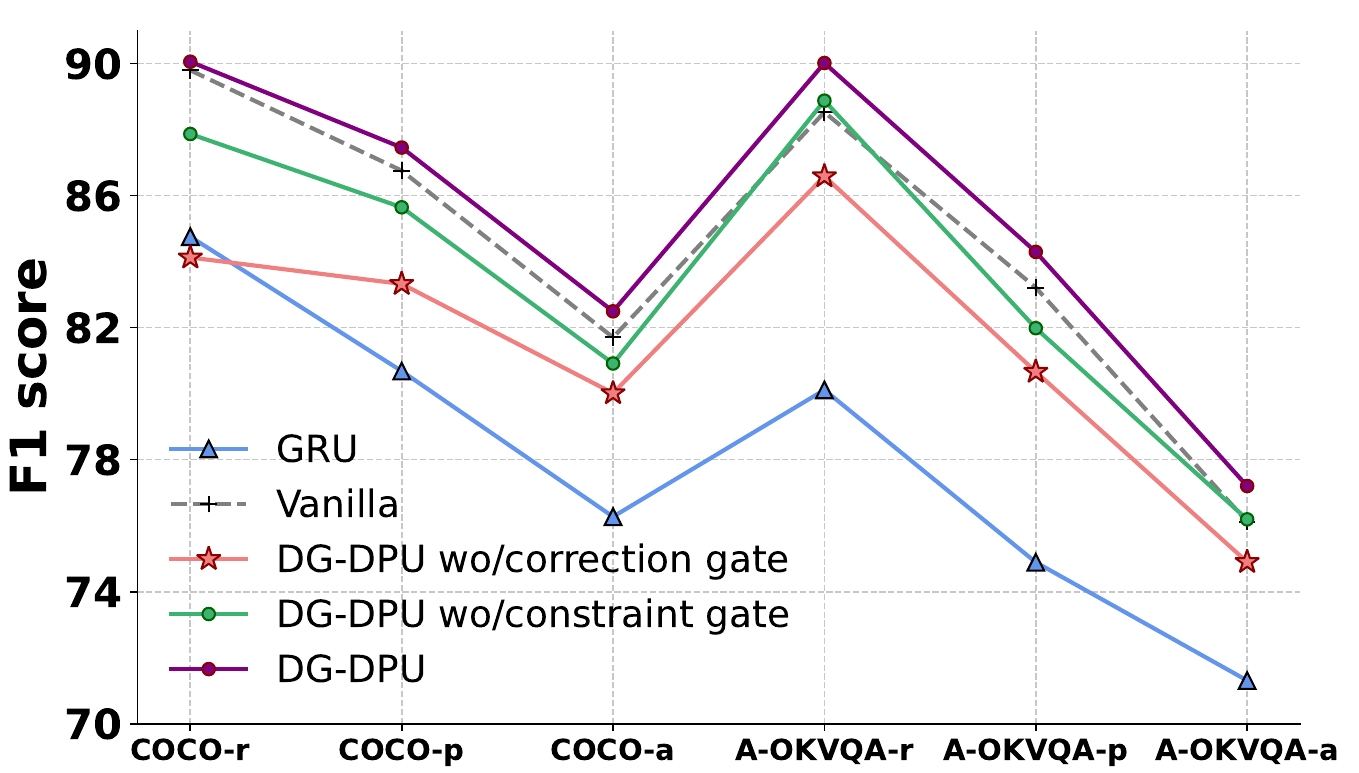}\subcaption{F1 score comparison}
    \end{minipage}
    \begin{minipage}{0.32\textwidth}
    \includegraphics[width=\textwidth]{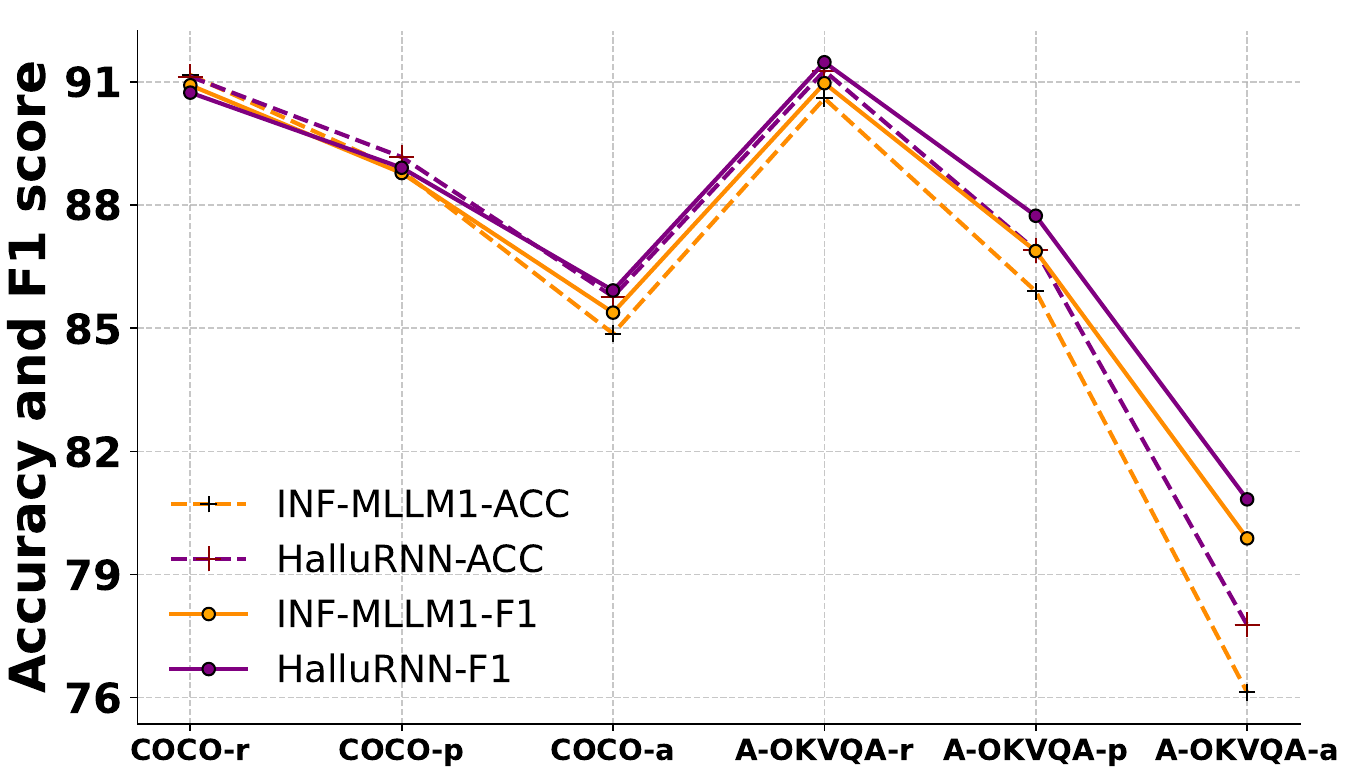}\subcaption{Acc. and F1 score comparison}
    \end{minipage}
    \caption{(a) Accuracy comparison across different configurations under the POPE benchmark using COCO and A-OKVQA datasets. The evaluated settings include our DG-DPU, as well as replacing the DG-DPU with a standard GRU module, Vanilla decoding, and ablations without the constraint gate or correction gate in the DG-DPU. $r$, $p$, and $a$ denote the random, popular, and adversarial settings, respectively. (b) F1 score comparison, using the same settings as in (a). (c) Accuracy and F1 score comparison between the Vanilla decoding and \ours, using the transferred DG-DPU trained on LLaVA-1.5. Results are reported on the INF-MLLM1 model under the POPE benchmark.  }\label{fig:ablation}
\end{figure*}

\paragraph{Benchmarks and Baselines}
We evaluate our proposed approach on multiple representative benchmarks designed to assess hallucinations in LVLMs, including POPE \citep{li2023evaluating} (utilizing the SEEM-annotated datasets from MSCOCO \citep{lin2014microsoft}, A-OKVQA \citep{schwenk2022okvqa} and GQA \citep{hudson2019gqa}), MM-Vet \citep{yu2023mm}, MMMU \citep{yue2024mmmu}, and CHAIR \citep{rohrbach2018object}. To ensure a comprehensive evaluation, we compare our method against a range of strong baselines, such as VCD \citep{leng2024mitigating}, VDD \citep{zhang2024debiasing}, DOLA \citep{chuang2023dola}, DAMO \citep{wangdamo2025}, POVID \citep{zhou2024aligning}, OPERA \citep{huang2024opera}, and DeCO \citep{wang2025mllm}, along with Vanilla LVLM outputs. For a fair comparison, the temperature hyperparameter is set to 0 for greedy search in all evaluations. Detailed descriptions of all benchmarks and baseline implementations are provided in the Appendix.

\subsection{Experimental Results}

\textbf{POPE} In the object hallucination benchmark POPE, We compare the performance of our method against other baselines on the POPE benchmark under three different settings across three datasets. As shown in the table \ref{Resul:POPE}, \ours achieves the highest accuracy and F1 scores across all datasets and sampling settings. Specifically, \ours surpasses challenging layer-wise decoding method DeCO by 2.54\%, 4.50\%, 5.47\% in accuracy improvement across three datasets in the popular setting. Compared to challenging contrastive decoding method VCD, \ours obtains performance improvements of 2.21\%, 2.25\% and 1.79\% in F1 score over the adversarial  setting. These results demonstrate that our decoding strategy effectively mitigates object hallucinations.

\textbf{MM-Vet} MM-Vet evaluates models across multiple vision-language capabilities, including recognition, OCR, knowledge, language generation, spatial reasoning, and math, using GPT-4 as the evaluator. As shown in Table~\ref{Resul:MM-Vet}, \ours outperforms most decoding strategies across all categories. Notably, in spatial awareness, \ours achieves 31.6\%, surpassing OPERA by 9.2\%. In the OCR category, \ours scores 27.1\%, significantly outperforming the layer-wise interaction method DAMO (22.4\%). These results highlight the robustness and generalizability of our method across diverse tasks, demonstrating its potential in addressing broader multimodal challenges.

\textbf{CHAIR} Table~\ref{Resul:CHAIR} presents results on the CHAIR benchmark for open-ended caption generation. \ours consistently outperforms other methods by significantly reducing object hallucinations. It achieves a 17.4\% and 3.0\% reduction in CHAIR\textsubscript{\textit{s}} and CHAIR\textsubscript{\textit{i}}, respectively, compared to the baseline. Notably, it surpasses the second-best method DeCO by 9.0\% in CHAIR\textsubscript{\textit{s}}, highlighting its effectiveness in addressing the gap between hallucinated and accurate tokens for reliable captioning.

\textbf{MMMU} To evaluate the generalization capability of \ours, we conduct experiments on the MMMU benchmark, which encompasses various disciplines and domains. The results, shown in Table \ref{Resul:mmmu}, indicate that \ours achieves the highest score among all baselines. Notably, \ours surpasses the layer-wise interaction method DeCO by 2.3\% and fine-tuning-based method POVID by 0.9\%, demonstrating  the strong generalization and adaptability of \ours across diverse tasks.

\subsection{Detailed Analysis}
\label{sec:ana}
We present a case study analyzing \ours effectiveness. As shown in Figure~\ref{fig:detailed_analysis}, when asked "Is there a tree in the image?" with a tree-free image, Vanilla decoding produces incorrect answers while \ours successfully corrects them. We analyze from two perspectives:
\ding{182} Token-level Prediction Dynamics: For key tokens such as the first token, the Vanilla model consistently assigns high probability to ``yes'', leading to hallucination. In contrast, \ours progressively adjusts this prediction in deeper layers, ultimately favoring ``no'' with high confidence. For less critical tokens like ``there'' and ``in'', \ours maintains stable and high probabilities, indicating that its focus remains on key semantic shifts.
\ding{183} Adaptive Information Flow via $g_e$: The gating scalar $g_e$ reveals the dynamic nature of information integration. Unlike fixed-weight methods such as DAMO, our approach adaptively determines how much historical information to retain at each layer. This dynamic gating highlights a key observation—information from earlier layers is essential for correcting hallucinations in later stages. Additional case studies are in Appendix.

\subsection{Ablation Studies}
\paragraph{Comparative Study with Standard RNN Variants}

To validate the effectiveness of our DG-DPU module within the \ours framework, we replace it with a standard GRU unit—chosen for its comparable two-input-one-output structure—and evaluate on the POPE benchmark under identical training settings. As shown in Figure \ref{fig:ablation} (a) and (b), DG-DPU consistently outperforms GRU, demonstrating its advantage in enforcing layer-wise consistency and mitigating hallucinations in LVLMs. This comparison highlights the strengths of DG-DPU in capturing the intricate dynamics between transformer layers, which are essential for maintaining stability and accuracy across deep networks. 

In contrast, the GRU fails to handle such complexities effectively, leading to poor performance, particularly in tasks requiring long-range dependencies and consistent representations across layers. Further details of the results, including comprehensive numerical comparisons, performance metrics, and a deeper analysis of the ablation study, are available in Appendix~\ref{sec:ablation_results}. These findings provide compelling evidence that the DG-DPU module significantly improves the stability and robustness of LVLMs, making it a crucial component for tasks that demand intricate understanding and consistent performance across deep model architectures.

\begin{table}[ht]
  \centering
  \caption{The detailed results on POPE benchmark using MSCOCO and A-OKVQA datasets when transferring \ours to INF-MLLM1}\adjustbox{width=0.48\textwidth}{
    \begin{tabular}{l|llcc}
    \hline
    \multicolumn{1}{l|}{Dataset} & \multicolumn{1}{l}{Setting} & Setting & \multicolumn{1}{p{4.96em}}{Accuracy} & \multicolumn{1}{p{4.415em}}{F1 Score} \bigstrut\\
    \hline
    \multirow{6}[6]{*}{MSCOCO} & \multicolumn{1}{l}{\multirow{2}[2]{*}{Random}} & Vanilla & 91.17 & 90.92 \bigstrut[t]\\
          &       & \ours  & 91.13 & 90.74 \bigstrut[b]\\
\cline{2-5}          & \multicolumn{1}{l}{\multirow{2}[2]{*}{Popular}} & Vanilla & 88.83 & 88.78 \bigstrut[t]\\
          &       & \ours  & 89.17 & 88.91 \bigstrut[b]\\
\cline{2-5}          & \multicolumn{1}{l}{\multirow{2}[2]{*}{Adversarial}} & Vanilla & 84.87 & 85.38 \bigstrut[t]\\
          &       & \ours  & 85.77 & 85.92 \bigstrut[b]\\
    \hline
    \multirow{6}[6]{*}{A-OKVQA} & \multicolumn{1}{l}{\multirow{2}[2]{*}{Random}} & Vanilla & 90.60  & 90.97 \bigstrut[t]\\
          &       & \ours  & 91.27 & 91.48 \bigstrut[b]\\
\cline{2-5}          & \multicolumn{1}{l}{\multirow{2}[2]{*}{Popular}} & Vanilla & 85.70  & 86.88 \bigstrut[t]\\
          &       & \ours  & 86.90  & 87.74 \bigstrut[b]\\
\cline{2-5}          & \multicolumn{1}{l}{\multirow{2}[2]{*}{Adversarial}} & Vanilla & 76.13 & 79.88 \bigstrut[t]\\
          &       & \ours  & 77.77 & 80.83 \bigstrut[b]\\
    \hline
    \end{tabular}}
  \label{tab:ablation_4}%
\end{table}%

\begin{table}[ht]
  \centering
  \caption{The detailed results on POPE benchmark using MSCOCO and A-OKVQA datasets when transferring \ours to Qwen-VL. $\dag$ denotes extra fine-tuning of DG-DPU in RefCOCO dataset.}\adjustbox{width=0.48\textwidth}{
    \begin{tabular}{l|llcc}
    \hline
    \multicolumn{1}{l|}{Dataset} & \multicolumn{1}{l}{Setting} & Setting & \multicolumn{1}{p{4.96em}}{Accuracy} & \multicolumn{1}{p{4.415em}}{F1 Score} \bigstrut\\
    \hline
    \multirow{6}[6]{*}{MSCOCO} & \multicolumn{1}{l}{\multirow{2}[2]{*}{Random}} & Vanilla &82.13 & 78.44\bigstrut[t]\\
          &       & \ours$\dag$  & 84.32& 79.20 \bigstrut[b]\\
\cline{2-5}          & \multicolumn{1}{l}{\multirow{2}[2]{*}{Popular}} & Vanilla & 81.93 & 78.25\bigstrut[t]\\
          &       & \ours$\dag$  & 82.02 & 79.33 \bigstrut[b]\\
\cline{2-5}          & \multicolumn{1}{l}{\multirow{2}[2]{*}{Adversarial}} & Vanilla & 80.97 & 77.37 \bigstrut[t]\\
          &       & \ours$\dag$  &82.38 &  79.94\bigstrut[b]\\
    \hline
    \multirow{6}[6]{*}{A-OKVQA} & \multicolumn{1}{l}{\multirow{2}[2]{*}{Random}} & Vanilla & 83.60  & 81.06 \bigstrut[t]\\
          &       & \ours$\dag$  & 85.40 & 83.45 \bigstrut[b]\\
\cline{2-5}          & \multicolumn{1}{l}{\multirow{2}[2]{*}{Popular}} & Vanilla & 83.47  & 80.94 \bigstrut[t]\\
          &       & \ours$\dag$  & 83.27 &79.32 \bigstrut[b]\\
\cline{2-5}          & \multicolumn{1}{l}{\multirow{2}[2]{*}{Adversarial}} & Vanilla & 78.97 & 76.95\bigstrut[t]\\
          &       & \ours$\dag$  & 79.10 &  77.72 \bigstrut[b]\\
    \hline
    \end{tabular}}
  \label{tab:ablation_5}%
\end{table}%

\paragraph{Assessing the Necessity of Constraint and Correction Gates in DG-DPU}
To evaluate the necessity of the dual-gate structure in DG-DPU, we conduct ablation studies by training two variants: one using only the Constraint Gate and the other using only the Correction Gate. Both variants are trained with the same hyperparameters as the full DG-DPU. We then evaluate these variants on the POPE benchmark. As shown in Figure \ref{fig:ablation} (a) and (b), the full DG-DPU outperforms both single-gate variants, demonstrating that the complementary roles of the two gates are crucial for effectively addressing hallucinations in LVLMs.
 
\paragraph{Cross-Model Transferability of \textbf{\ours}}
To evaluate the robustness and generalizability of \ours, we apply it to a another LVLM, INF-MLLM1 \citep{zhou2023infmllm}. We transfer the DG-DPU module trained on LLaVA-1.5 and integrate it into INF-MLLM1 without extra fine-tuning, maintaining the original \ours structure. As shown in Figure \ref{fig:ablation} (c), evaluation on the POPE benchmark shows that \ours consistently reduces hallucinations, proving its effectiveness as a plug-and-play solution for enhancing layer-wise consistency across LVLM architectures. Table \ref{tab:ablation_4} represents the results when transferring \ours trained on LLaVA to INF-MLLM1.

Furthermore, we performed additional experiments by applying it to Qwen-VL~\cite{bai2023qwenvlversatilevisionlanguagemodel}. The DG-DPU module was trained on Qwen-VL with 8.8k val dataset form RefCOCO~\cite{kazemzadeh-etal-2014-referitgame} with default parameter settings in Qwen-VL. The results, summarized in Table \ref{tab:ablation_5}, show that \ours again delivers consistent improvements in both accuracy and F1 score across the MSCOCO and A-OKVQA datasets, particularly in the Adversarial setting, further validating its versatility and robustness when transferred to different model structures. These findings highlight the potential of \ours in enhancing the performance and stability of various LVLMs.

\section{Conclusion}
In this paper, we introduce \ours, a lightweight and generalized architecture-level approach designed for mitigating hallucinations in Large Vision-Language Models (LVLMs). By introducing Dual-Gate Depth Propagation Unit (DG-DPU), a novel recurrent module shared across Transformer layers, our \ours enables adaptive cross-layer reasoning to address representation drift and reasoning inconsistencies during inference. Extensive evaluations on various benchmarks show that \ours consistently addresses hallucinations compared to other challenging methods. Furthermore, systematic comparisons with standard RNN variants such as GRU highlight the effectiveness of our specialized design in stabilizing hidden state propagation across layers.

\section{Impact Statement.}
HalluRNN introduces a lightweight recurrent architecture that improves stability and reduces hallucinations in long-horizon sequential reasoning. By enhancing reliability while maintaining computational efficiency, it enables the deployment of robust models in real-time, resource-constrained environments. Our work highlights the importance of addressing biases and failure modes, particularly in safety-critical applications. This research contributes to the development of more reliable, sustainable, and ethically deployed AI systems.

\bibliography{reference}
\bibliographystyle{icml2026}

\clearpage

\input{appendix}

\end{document}

%% file: appendix.tex
\appendix
\onecolumn

\section{More Analysis about Addressing Hallucinations in LVLMs}\label{apen:rela}
With the success of DoLA \cite{chuang2023dola} in LLMs, contrastive decoding (CD) variants have emerged to tackle hallucinations in LVLMs. VCD \citep{leng2024mitigating} addresses object hallucinations by contrasting output logits from Vanilla with those obtained from distorted images. VDD \citep{zhang2024debiasing} introduces debiasing strategies, including post-hoc debiasing and debias sampling, to tackle biases in LVLM-generated content. HIO \citep{chen2024alleviating} enhances the distinction between hallucinatory and target tokens using a fine-tuned theoretical preference model, enabling more effective contrastive decoding. IBD \citep{zhu2024ibd} mitigates hallucinations by contrasting predictions from a standard LVLM with those from an image-biased LVLM, reinforcing information closely aligned with image content. CODE \citep{kim2024code} utilizes self-generated description as contrasting visual counterpart to correct hallucinatory responses. SID \citep{huo2025selfintrospective} adaptively amplifies and then subtracts vision-and-text association hallucinations, based on findings that pre-trained LVLMs can introspectively evaluate the importance of visual information using preceding tokens. OPERA \citep{huang2024opera} introduces a penalty term on model logits during beam search decoding to mitigate over-reliance on hallucinated information. DeGF \citep{zhang2025self} uses feedback from text-to-image generative models to mitigate hallucinations in LVLMs by generating an image from the initial response for verification and correction.

\section{More Training Details} \label{appendix:training_details}
POVID \citep{zhou2024aligning} originally constructed their training dataset using a novel Direct Preference Optimization (DPO) approach, resulting in paired samples of preferred and hallucinated responses. In our case, since we do not require the hallucinated responses, we only utilize the preferred outputs — that is, the original answers from the 17K examples. The learning rate is set to 1e-5, and the batch size is 10. All other hyperparameters follow the official LLaVA fine-tuning setup.

\section{More Details about Benchmarks and Baselines} \label{appendix:benchmark_model_details}

We evaluate our proposed methods on various benchmarks designed to assess model hallucinations in LVLMs. (1) MME \citep{fu2023mme} evaluates both the perception and cognition abilities of LVLMs across 14 subtasks, employing manually constructed instruction-answer pairs to ensure robustness and comparability. (2) POPE \citep{li2023evaluating} is a novel framework specifically designed to systematically assess object hallucination in LVLMs, utilizing the SEEM-annotated datasets from MSCOCO \citep{lin2014microsoft}, A-OKVQA \citep{schwenk2022okvqa} and GQA \citep{hudson2019gqa}. (3) MM-Vet \citep{yu2023mm} focuses on evaluating LVLMs on complex tasks by examining their ability to integrate six core vision-language capabilities. (4) MMMU \citep{yue2024mmmu} is a large-scale benchmark that assesses LVLMs on expert-level tasks across 30 subjects and 183 subfields, emphasizing advanced reasoning and domain-specific knowledge. (5) CHAIR \citep{rohrbach2018object} is a specialized tool for evaluating object hallucination in image captioning tasks. We use the same subset of CHAIR as in \citep{wang2025mllm}.

\begin{figure*}
	\centering
    \vspace{-1pt}
	\begin{minipage}{0.36\textwidth}
	{ 			\includegraphics[width=\textwidth]{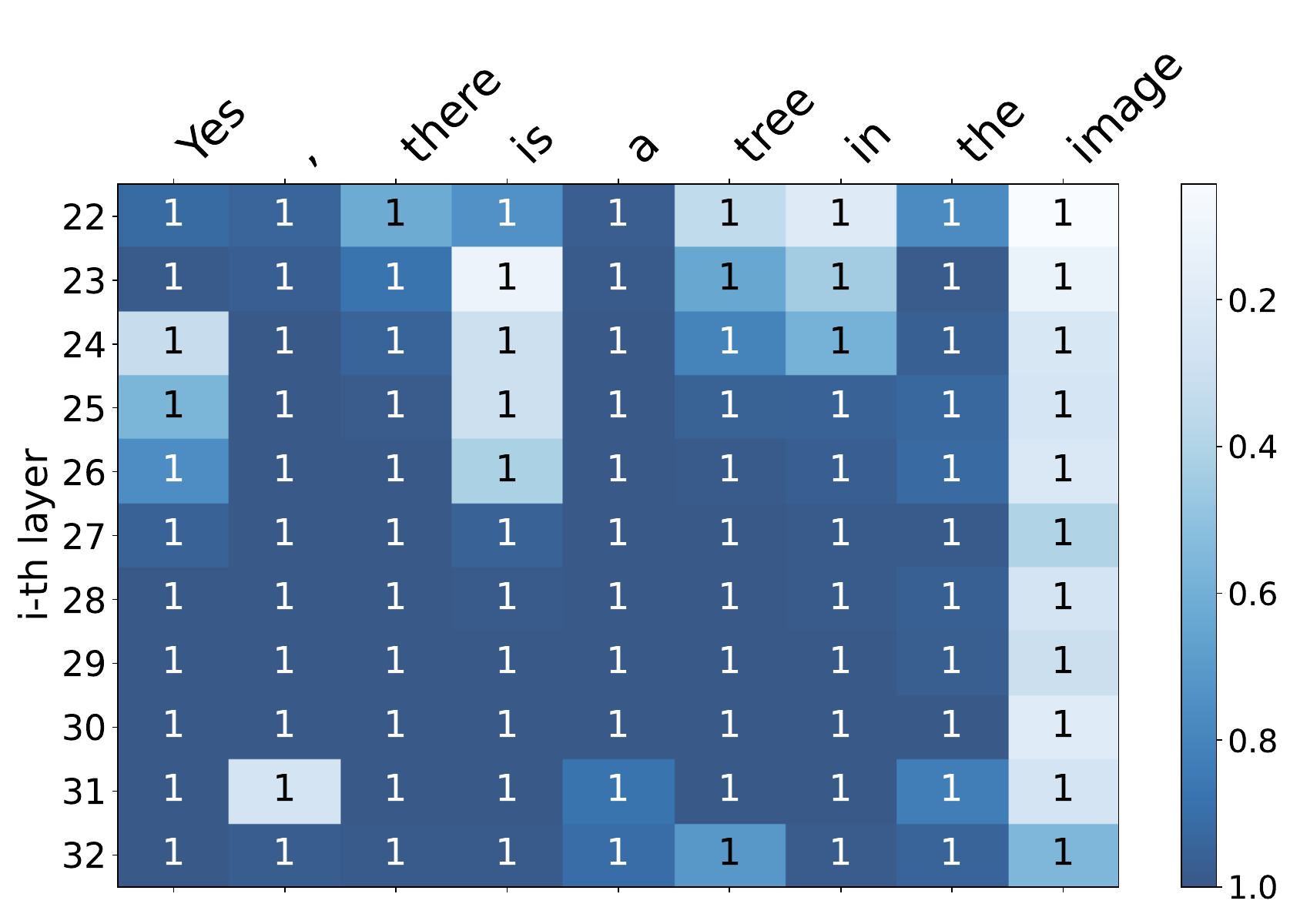}}\subcaption{Vanilla}
	\end{minipage}
	\begin{minipage}{0.36\textwidth}
	{ 			\includegraphics[width=\textwidth]{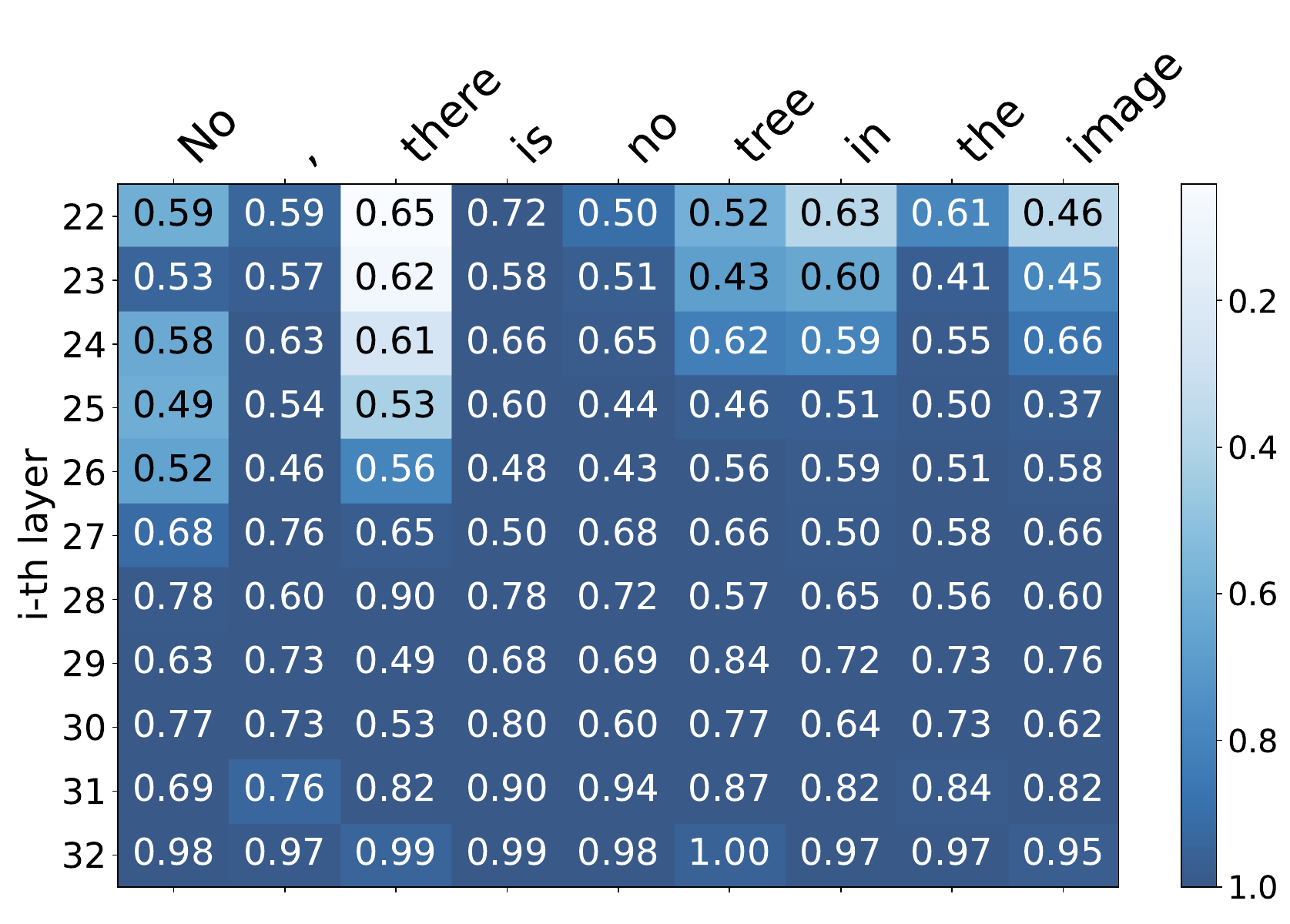}}\subcaption{\ours}
	\end{minipage}
	\begin{minipage}{0.25\textwidth}
	{ 			\includegraphics[width=\textwidth]{Figures/1/8_COCO_val2014_000000560474.jpg}}
	\end{minipage}
	\centering
    \vspace{-1pt}
	\begin{minipage}{0.36\textwidth}
	{ 			\includegraphics[width=\textwidth]{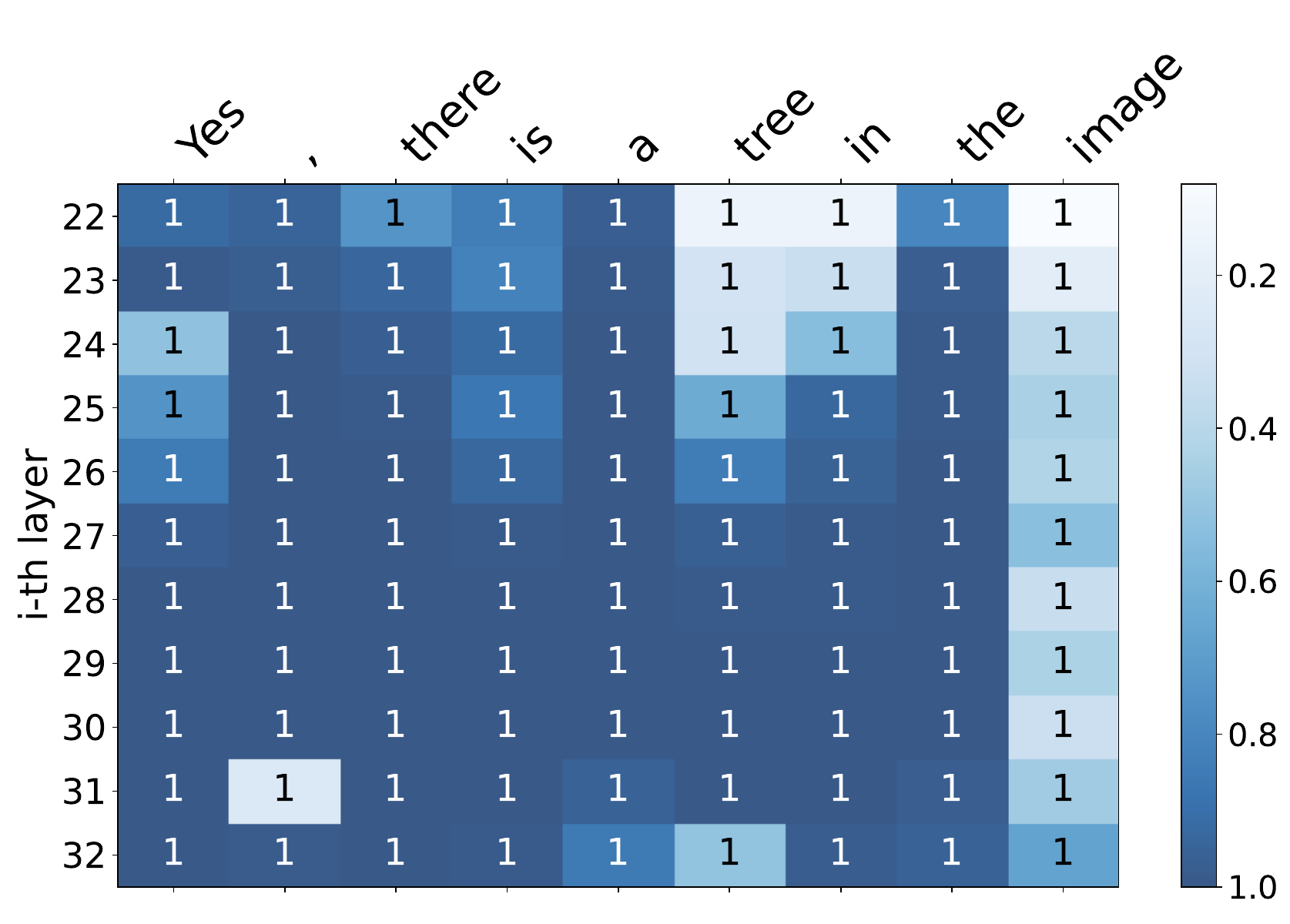}}\subcaption{Vanilla}
	\end{minipage}
	\begin{minipage}{0.36\textwidth}
	{ 			\includegraphics[width=\textwidth]{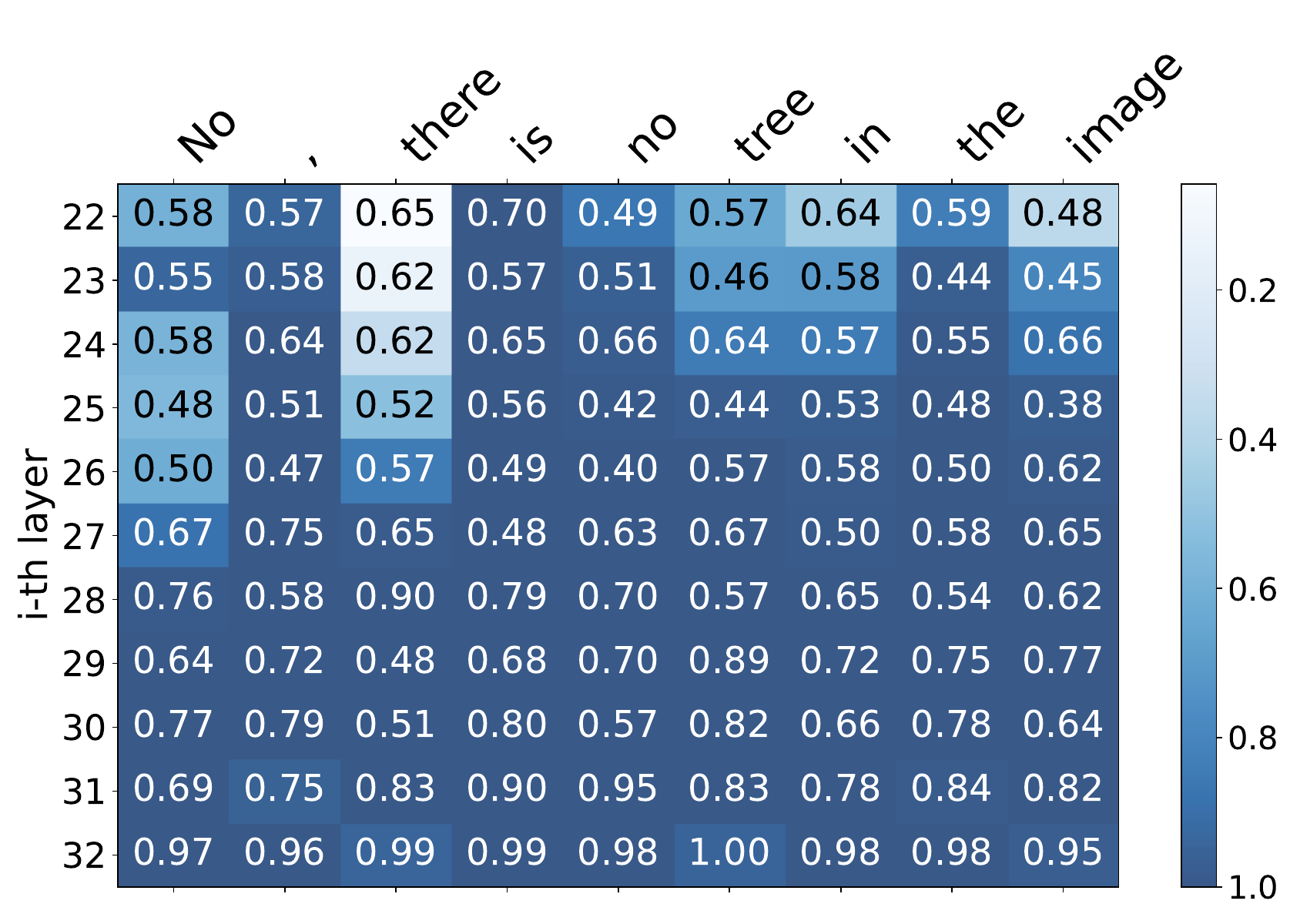}}\subcaption{\ours}
	\end{minipage}
	\begin{minipage}{0.25\textwidth}
	{ 			\includegraphics[width=\textwidth]{Figures/1/9_COCO_val2014_000000579970.jpg}}
	\end{minipage}    \caption{Two more case studies: The visualization compares the Vanilla and \ours decoding methods in addressing the question posed by the textual input \textit{``Is there a tree in the image?''}, alongside the corresponding image from the COCO 2014 dataset. The color density of each grid indicates the probability of the output word at the $i$-th layer, while the displayed values represent the $g_e$ values, which show a clear upward trend as the network depth increases.}
    \vspace{-2mm}
\label{fig:two_more_case_study}
\end{figure*}

We compared our method with several baselines. Visual Contrastive Decoding (VCD) \citep{leng2024mitigating} mitigates object hallucinations by contrasting output distributions derived from original and distorted visual inputs. VDD \citep{zhang2024debiasing} introduces use post-hoc debiasing and debias sampling to tackle biases in LVLM-generated content for mitagating hallucinations. DOLA \citep{chuang2023dola} addresses hallucinations in LLMs by contrasting logits from later and earlier Transformer layers, and we transfer DOLA for use with LVLMs. DAMO \citep{wangdamo2025} introduces momentum decoding to enforce consistency across layers, reducing hallucinations in LVLMs. POVID \citep{zhou2024aligning} tackles hallucinations through an RLHF pipeline via Direct Preference Optimization, with results reported using their released model weights. OPERA \citep{huang2024opera} reduces hallucinations by applying an Over-trust Penalty and a Retrospection-Allocation strategy. DeCO \citep{wang2025mllm} adaptively selects relevant preceding layers and integrates them into the final layer to enhance layer interaction to reduce hallucinations.

\section{Supplementary Results on the MME Benchmark} 
We evaluate the performance of our proposed \ours on the comprehensive MME benchmark. As shown in Table~\ref{Resul:MME_percep}, \ours achieves a score of 1511.35, ranking second overall and outperforming all baselines except DAMO. Notably, \ours surpasses Vanilla decoding by 20.66 points. Compared to other layer-wise approaches, \ours outperforms DOLA and DeCO by 20.74 and 51.06 points, respectively, demonstrating the effectiveness of our recurrent cross-layer reasoning design.
\begin{table*}[ht]
    \centering    
    \small   
    \vspace{-3.5mm}
    \caption{Experimental results of various decoding strategies on MME benchmark in the perception category utilizing LLaVA-1.5 model.}
    \adjustbox{width=0.9\textwidth}{
    \begin{tabular}{ccccccccccccc}
        \toprule
         Decoding & Exist. & Count & Pos. & Color & Poster & Celeb. & Scene & Land. & Art & Ocr& Total\\ \midrule
        Vanilla & 190.00 & 160.00 & 138.30 & 165.00 & 140.48 & 135.88 & 156.25 & 161.25 & 118.50 & 125.00 & 1490.69 \\
                 VCD & 180.00 & 148.30 & 135.00 & 170.00 & 141.84 & 144.71 & 151.75 & 167.50 & 123.75 & 110.00 & 1472.88\\

                 VDD & 180.00 & 150.00 & 138.33 & 165.00 & 135.37 & 137.35 & 154.00 & 169.00 & 124.00 & 122.50 & 1475.56 \\
                
                 DOLA&190.00&158.33&143.33&165.00&139.46&133.24&157.00&160.50&118.75&125.00&1490.61\\
                POVID&190.00&155.00&143.33&165.00&142.52&136.76&157.00&162.75&118.25&132.50&1503.12\\

                DAMO&195.00 &150.00 &143.33 &165.00 &144.56 &134.12 &156.25 &166.00 &120.00 &140.00&\textbf{1514.26}\\

                OPERA& 195.00  &158.33 &143.33 &175.00 &142.52 &130.00 &158.50 &159.00 &116.50 &117.50 & 1495.68\\

                DeCO& 190.00 & 148.33 & 115.00 & 165.00 & 149.66 & 135.29 & 152.25 & 164.50 & 110.25 & 130.00& 1460.29\\
                \ours & 190.00 & 156.67 & 148.33 & 170.00 & 138.44 & 134.41 & 156.25 & 159.00 & 118.25 & 140.00 & 1511.35 \\
                \bottomrule                
    \end{tabular}
    }\label{Resul:MME_percep}
\end{table*}

\section{Limitations}\label{sec:limitation}
While \ours demonstrates strong performance and generalization across multiple benchmarks, it still inherits a key limitation from existing LVLMs: dependence on the quality of intermediate representations. If early-layer representations are significantly corrupted or misleading (e.g., due to severe visual noise or adversarial input), the recurrent module may propagate errors along the network depth. Addressing this issue may require integrating robustness-aware mechanisms or confidence estimation into the cross-layer reasoning process.

\begin{table}[ht]
  \centering
  \caption{The detailed results on POPE benchmark using MSCOCO and A-OKVQA datasets when replacing DG-DPU with standard GRU in \ours.}
    \begin{tabular}{l|lcc}
    \hline
    \multicolumn{1}{l|}{Datasets} & \multicolumn{1}{p{4.835em}}{Setting} & \multicolumn{1}{l}{Accuracy} & \multicolumn{1}{l}{F1 Score} \bigstrut\\
    \hline
    \multirow{3}[2]{*}{MSCOCO} & Random & 82.57 & 84.75 \bigstrut[t]\\
          & Popular & 76.80  & 80.68 \\
          & Adversarial & 70.60  & 76.27 \bigstrut[b]\\
    \hline
    \multirow{3}[2]{*}{A-OKVQA} & Random & 75.23 & 80.11 \bigstrut[t]\\
          & Popular & 66.57 & 74.89 \\
          & Adversarial & 59.87 & 71.31 \bigstrut[b]\\
    \hline
    \end{tabular}%
  \label{tab:ablation_1}%
\end{table}%

\begin{table}[ht]
  \centering
  \caption{The detailed results on POPE benchmark using MSCOCO and A-OKVQA datasets when only the Constraint Gate is retained in DG-DPU.}
    \begin{tabular}{l|lcc}
    \hline
    \multicolumn{1}{l|}{Datasets} & \multicolumn{1}{p{4.835em}}{Setting} & \multicolumn{1}{l}{Accuracy} & \multicolumn{1}{l}{F1 Score} \bigstrut\\
    \hline
    \multirow{3}[2]{*}{MSCOCO} & Random & 86.17 & 84.12 \bigstrut[t]\\
          & Popular & 85.33  & 83.32 \\
          & Adversarial & 77.69  & 80.01 \bigstrut[b]\\
    \hline
    \multirow{3}[2]{*}{A-OKVQA} & Random & 87.33 & 86.58 \bigstrut[t]\\
          & Popular & 79.26 & 80.66 \\
          & Adversarial & 70.84 & 74.90 \bigstrut[b]\\
    \hline
    \end{tabular}%
  \label{tab:ablation_2}%
\end{table}%

\begin{table}[ht]
  \centering
  \caption{The detailed results on POPE benchmark using MSCOCO and A-OKVQA datasets when only the Correction Gate is retained in DG-DPU.}
    \begin{tabular}{l|lcc}
    \hline
    \multicolumn{1}{l|}{Datasets} & \multicolumn{1}{p{4.835em}}{Setting} & \multicolumn{1}{l}{Accuracy} & \multicolumn{1}{l}{F1 Score} \bigstrut\\
    \hline
    \multirow{3}[2]{*}{MSCOCO} & Random & 83.37 & 87.86 \bigstrut[t]\\
          & Popular & 85.97  & 85.64 \\
          & Adversarial & 80.63  & 80.91 \bigstrut[b]\\
    \hline
    \multirow{3}[2]{*}{A-OKVQA} & Random & 88.53 & 88.87 \bigstrut[t]\\
          & Popular & 80.63 & 81.98 \\
          & Adversarial & 70.23 & 76.19 \bigstrut[b]\\
    \hline
    \end{tabular}%
  \label{tab:ablation_3}%
\end{table}%

\section{Additional Case Studies}\label{sec:more_case_studies}
We present two additional case studies shown in Figure \ref{fig:two_more_case_study}, which illustrate the detailed information flow between Vanilla and \ours. Both examples are posed the same textual question, ``Is there a tree in the image?'' alongside the corresponding image in each raw. Specifically, these two examples highlight the same finds we discussed earlier.

\section{Comparative Study with Standard RNN Variants}\label{sec:ablation_results}

Table \ref{tab:ablation_1}, Table \ref{tab:ablation_2}, Table \ref{tab:ablation_3} represents the performance on POPE benchmark when replacing DG-DPU with standard GRU within \ours, only retaining Constraint Gate, only retaining Correction Gate, respectively.

\section{Licenses and copyrights across assets}
\label{appendix: license}

\begin{enumerate}

    \item \textbf{MME}
    \begin{itemize}
        \item Citation: \citet{fu2023mme}
        \item Asset Link:\url{[https://github.com/BradyFU/Awesome-Multimodal-Large-Language-Models/tree/Evaluation]}
    \end{itemize}
    
    \item \textbf{POPE}
    \begin{itemize}
        \item Citation: \citet{li2023evaluating}
        \item Asset Link:\url{[https://github.com/RUCAIBox/POPE]}
        \item License: \url{[https://github.com/RUCAIBox/POPE/blob/main/LICENSE]}
    \end{itemize}
        
    \item \textbf{MM-Vet}
    \begin{itemize}
        \item Citation: \citet{yu2023mm}
        \item Asset Link: \url{[https://github.com/yuweihao/MM-Vet]}
        \item License: \url{[https://github.com/yuweihao/MM-Vet/blob/main/LICENSE]}
    \end{itemize}

    \item \textbf{CHAIR}
    \begin{itemize}
        \item Citation: \citet{rohrbach2018object}
        \item Asset Link: \url{[https://github.com/Maxlinn/CHAIR-metric-standalone]}
    \end{itemize}

    \item \textbf{MMMU}
    \begin{itemize}
        \item Citation: \citet{yue2024mmmu}
        \item Asset Link: \url{[https://huggingface.co/datasets/MMMU/MMMU]}
        \item License: \url{[https://choosealicense.com/licenses/apache-2.0/]}
    \end{itemize}
    
\end{enumerate}

\section{Broader Impacts}\label{sec:broader_impacts}
This work enhances the stability and reliability of Large Vision-Language Models (LVLMs), which holds potential for positive societal impact in areas such as education, accessibility, and human-AI collaboration. By mitigating hallucinations, our method can help reduce misinformation and improve user trust in multimodal AI systems. However, the improved fluency and coherence of outputs may also inadvertently increase the risk of generating persuasive yet inaccurate or harmful content, especially if deployed without adequate safeguards. Therefore, while the technique is promising, responsible deployment and further research into alignment and safety remain crucial.
